%% file: main.tex
\documentclass[10pt,twocolumn,letterpaper]{article}

\usepackage[pagenumbers]{iccv} 
\usepackage{tikz}
\usepackage{amsmath,amsfonts,amssymb, mathrsfs, extarrows,multirow,xcolor,subfiles,xr, tikz, balance}
\usepackage[accsupp]{axessibility}

\definecolor{cvprblue}{rgb}{0.21,0.49,0.74}
\definecolor{iccvblue}{rgb}{0.21,0.49,0.74}
\definecolor{lightred}{rgb}{1,0.4,0.4}
\definecolor{lightyellow}{rgb}{1,1,0.4}
\definecolor{lightorange}{rgb}{1,0.647,0.4}

\newcommand{\changes}[1]{\textcolor{black}{#1}}

\usepackage[pagebackref,breaklinks,colorlinks,allcolors=cvprblue]{hyperref}
\def\paperID{2} 
\def\confName{ICCV}
\def\confYear{2025}

\title{Human Vision Constrained Super-Resolution}

\author{
Volodymyr Karpenko \qquad
Taimoor Tariq \qquad
Jorge Condor \qquad
Piotr Didyk \\[4pt]
{\tt\small \{volodymyr.karpenko,taimoor.tariq,jorge.condor,piotr.didyk\}@usi.ch} \\[6pt]
Università della Svizzera Italiana (USI), Lugano, Switzerland
}

\begin{document}
\maketitle
\begin{figure*}
    \centering
    \includegraphics[width=\linewidth]{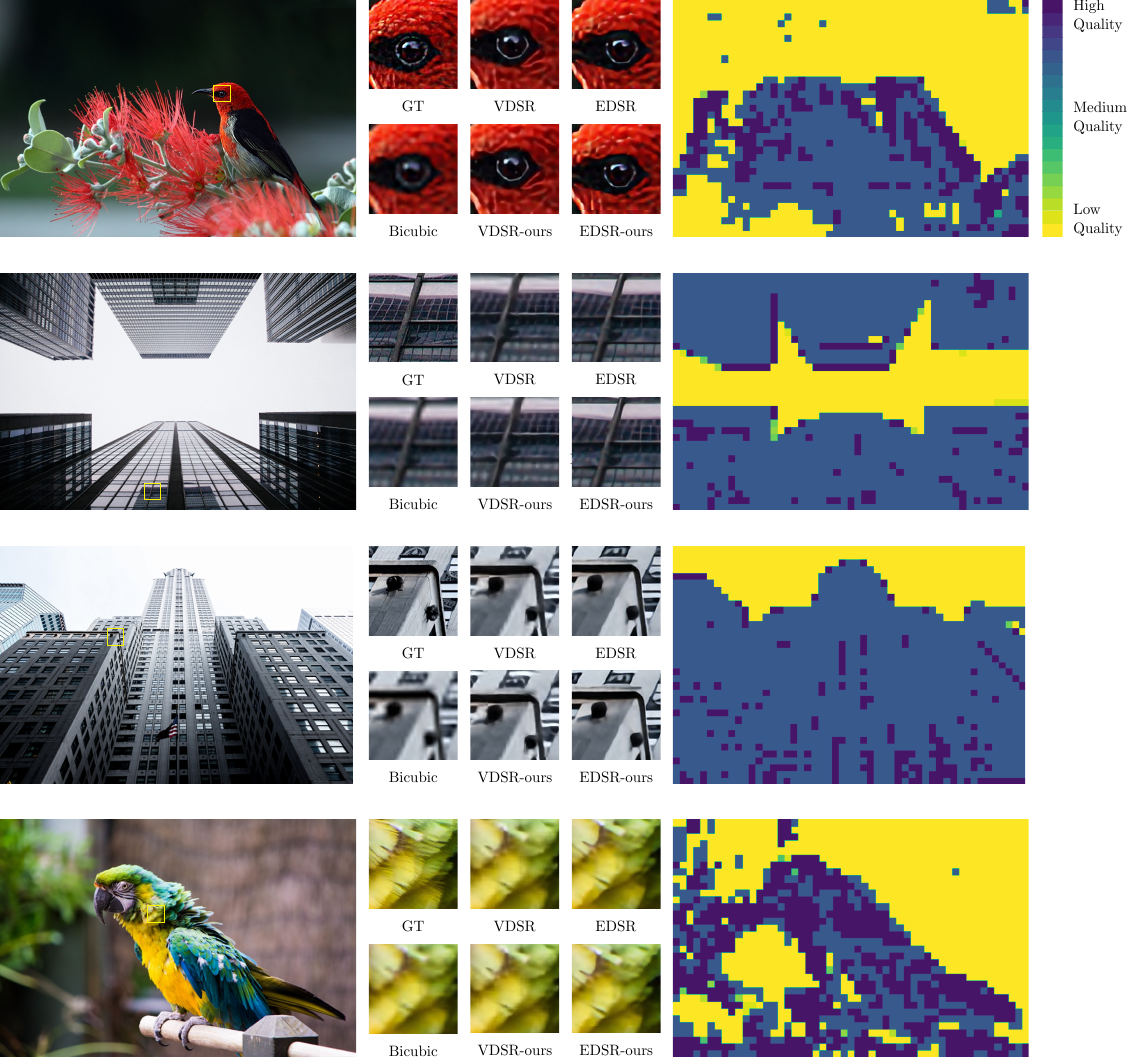}
    \caption{Visual results of our method compared to the original networks. On the right, we can observe the maps produced by our perceptual model.
    }
    \label{fig:visual_results}
\end{figure*}
\input{sec/00_definitions}
\input{sec/0_abstract}    
\input{sec/1_intro}
\input{sec/2_related_work}
\input{sec/2.5_background}
\input{sec/3_methods}
\input{sec/4_results}
\input{sec/4.5_future_work}
\input{sec/5_conclusion}

\section*{Acknowledgement}
This project has received funding from the European Research Council (ERC) under the European Union's Horizon 2020 research and innovation program (grand agreement No 804226).

{
    \small
    \bibliographystyle{ieeenat_fullname}
    \bibliography{main}
}
\subfile{sec/X_suppl}

\end{document}

%% file: sec/00_definitions.tex
\newcommand{\img}{\mathcal{I}}
\newcommand{\R}{\mathbb{R}}

%% file: sec/0_abstract.tex
\begin{abstract}
Modern deep-learning super-resolution (SR) techniques process images and videos independently of the underlying content and viewing conditions. 
However, the sensitivity of the human visual system (HVS) to image details changes depending on the underlying image characteristics, such as spatial frequency, luminance, color, contrast, or motion; as well viewing condition aspects such as ambient lighting and distance to the display. 
This observation suggests that computational resources spent on up-sampling images/videos may be wasted whenever a viewer cannot resolve the synthesized details i.e the resolution of details exceeds the resolving capability of human vision. 
Motivated by this observation, we propose a human vision inspired and architecture-agnostic approach for controlling SR techniques to deliver visually optimal results while limiting computational complexity. 
Its core is an explicit Human Visual Processing Framework (HVPF) that dynamically and locally guides SR methods according to human sensitivity to specific image details and viewing conditions. 
We demonstrate the application of our framework in combination with network branching to improve the computational efficiency of SR methods. 
Quantitative and qualitative evaluations, including user studies, demonstrate the effectiveness of our approach in reducing FLOPS by factors of 2$\times$ and greater, without sacrificing perceived quality. The source code is available at \url{https://github.com/karpev98/hvc-sr}.
\end{abstract}

%% file: sec/1_intro.tex
\section{Introduction}\label{sec:introduction}
Super-resolution (SR) has quickly become a fundamental tool in imaging and media distribution, given the increasing requirements of delivering higher quality content at lower bandwidths, and as general compression tools to deal with escalating imaging sensor resolutions. 
In media production such as virtual reality, augmented reality, and video games, SR is indispensable to cater to the high-resolution, high-framerate requirements of modern displays and low power budgets~\cite{foveated_nvidia16, nvidia_dlss, amd_fsr, intel_xess}. 
With the advent of hardware-specific accelerators for efficiently running DL models~\cite{nvidia_tensor_core}, most modern approaches, even for real-time needs, involve using convolutional neural networks trained on large priors of natural images, which during run-time take low-resolution images as input and produce higher resolution versions. 
However, even with ever-increasing computational power at our disposal, the computational burden of high-quality SR methods is still problematic~\cite{adobe_lightroom_super_resolution}. 
In fact, state-of-the-art methods on real-time SR using neural networks, such as Bicubic++~\cite{bilecen2023bicubic}, directly compete with efficient, classic bicubic interpolation techniques, yielding marginal quantitative improvements while still struggling to compete in runtime.
At the same time, the human visual system (HVS) is compressive by nature~\cite{zhaoping2006theoretical}, meaning it has limited capabilities to resolve detail beyond some thresholds determined by viewing conditions, spatial frequencies, color or motion; any further improvement in reconstruction quality achieved by using a more expensive model can be seen as wasted resources. 
Our key insight is that, given that SR images are to be observed by a human, we can take advantage of these naturally compressive capabilities of the HVS~\cite{zhaoping2006theoretical} to process differently areas of the input image depending on its characteristics and human sensitivity to those, such as spatial frequency, luminance, color, contrast, or even, in the case of videos, motion. 
Computational resources are allocated to perceptually meaningful areas, as determined by our low-level visual model; analogously to how lossy compression schemes such as JPEG similarly allocate memory by leveraging the HVS. 
\newline\indent This observation is the foundation of our work. 
First, we quantify the frequency reconstruction capability of baseline SR neural networks after each layer, doing so through attenuation curves. 
Then, after dividing an image or individual frame into a set of patches, our Human Visual Processing Framework (HVPF) (grounded on recent contrast-sensitivity functions~\cite{stelacsf}) predicts how many layers of the network should the patch be processed through; visually sensitive patches will be processed by the whole model, whereas the least visually meaningful content will simply resort to bicubic interpolation, \cref{fig:visual_results} shows a visual example. 
We apply our model to other use cases (with the goal of computational efficiency) such as selecting a network of appropriate depth from a set of candidates. 
In all of these cases, the goal is make the inference as fast as possible without noticeable quality degradation.
Through a series of user studies and quantitative measurements of quality and runtime, we demonstrate that our HVPF enables faster runtime with no perceivable loss of quality when compared with the baseline models. Furthermore, contemporary SR methods often only consider foveal vision, making them suboptimal for wide-field-of view systems such as AR/VR displays. We also propose an eccentricity-aware extension of our model for VR and AR applications, where computational efforts are further directed as per the loss of human visual acuity in peripheral vision~\cite{foveated_nvidia16}. Our code and pre-trained models are publicly available at \url{https://github.com/karpev98/hvc-sr}.

%% file: sec/2_related_work.tex
\section{Related Work}\label{sec:related_work}
\paragraph{Single-frame SR}
Super-Resolution or image-upsampling has traditionally been addressed through interpolation-based techniques. Nearest-Neighbor, Bilinear or Bicubic interpolation techniques are commonly used, featuring increasingly bigger receptive fields and perceived quality, at the cost of performance. However, interpolation-based techniques that solely consider the input image are fundamentally limited in terms of upsampling capacity, as high frequency signals lost in the downsampled or compressed image could never be recovered from the signal itself. Leveraging natural image statistics, either explicitly~\cite{yang08} or implicitly~\cite{Wang2021UnsupervisedDR} has shown greater potential, as lost frequencies can be composed from the expected distribution already present in natural images. Implicit (Deep-Learning based) methods have recently received the greatest attention, as large datasets of high resolution images~\cite{imagenet,laion5b} and advances in neural-network architectures~\cite{alexnet,He2015DeepRL,Dosovitskiy2020AnII}. 
\newline\indent Early efforts were based on convolutional neural networks~\cite{Dong2014ImageSU,kim16,Wang2018AFP}. In practice, they learned sets of convolutional filters adapted to different features, enabling them to better reconstruct missing frequencies in a content-aware manner. These were usually trained end-to-end via downsampling HR images and upsampling the result to recover the original signal, usually employing pixel-error or perceptual~\cite{Zhang2018TheUE} loss functions. A popular extension of these approaches leveraged adversarial training~\cite{goodfellow14} with latent CNNs for higher quality results~\cite{Wang2018ESRGANES}. More recently, some methods rely on Vision Transformers~\cite{Dosovitskiy2020AnII, Liang2021SwinIRIR, Baghel2023SRTransGANIS, Lu2021TransformerFS} for their latent architecture. The self-attention mechanism inherent in the architecture enables capturing long-range relationships within the image content, as opposed to the solely local receptive fields of CNN-based approaches. State-of-the-art methods nowadays rely on diffusion~\cite{rombach2021highresolution} rather than adversarial training~\cite{Li2021SRDiffSI, Xia2023DiffIRED}, featuring improved learning stability and quality. Despite their quality however, diffusion methods (either with vision transformer or CNN backbones) are computationally expensive and usually disregarded in applications where performance is primed.

\paragraph{Temporally-consistent Video SR}
Temporally-consistent video SR has received ample attention in the literature, for both pre-rendered video~\cite{Li2023TowardsHA} and real-time content (i.e. videogames)~\cite{nvidia_dlss,intel_xess,amd_fsr}. The main difference between single image SR is the availability of additional frames, as well as extra information from the rendering engine in the case of videogames (i.e. motion vectors, material information). Traditionally, single-image learning-based approaches struggle to provide consistent SR across frames due to the implicit, difficult-to-interpret latent space they leverage, which does not guarantee temporally consistent outputs. To ensure smoothness across frames, most works simply condition the current frame upsampling on previous or subsequent frames through motion vectors or optical flow to ensure smoothness~\cite{nvidia_dlss}. 
\newline\indent There is a fundamental difference between the traditional approach to video SR and our proposal that is important to clarify. Traditionally, video SR methods aim to achieve a better spatial reconstruction due to the availability of multiple frames. However, human sensitivity to spatial details decreases with movement, so a human vision centric approach should leverage this and reduce quality as a function of movement magnitude, without noticeable visual degradation. Our proposed model quantifies the loss of visual acuity with motion, and appropriately decreases the spatial quality of SR. Therefore, we can get faster per-frame SR with videos, in a manner adaptive to factors such as the nature of the content and amount of movement. 

%% file: sec/2.5_background.tex
\section{Background}\label{sec:background}

\subsection{Low-Level Human Vision}\label{sec:hv}
Owing to evolutionary imperatives of efficiency, the HVS has evolved to be a compressive system~\cite{zhaoping2006theoretical}. 
What this essentially means is that we are not able to resolve all the visual information that is coming into our eyes. 
The first critical aspect is that due to center surround receptive field of the early visual system, we are much more sensitive to variations in contrast rather than absolute luminance \cite{Shapley1993ContrastSA}.
Furthermore, it is well known that human contrast perception is highly dependent on factors such as spatial frequency and luminance. Due to the nature of the collective receptive field our early human vision, we are most sensitive to a narrow band of spatial frequencies, and our sensitivity falls of at lower and higher spatial frequencies. 
This phenomena is aptly captured by a model of the early visual system called the Contrast Sensitivity Function (CSF) \cite{Barten2003FormulaFT}. 
The CSF is highly dependent on factors such as local adapting luminance, size of the stimulus, eccentricity, and the amount of movement. For example, universal sensory models such as the Weber-Fechner law tell us that our ability to detect contrast decreases with increasing luminance \cite{Mantiuk2011HDRVDP2AC}. 
Furthermore, due to effects such as contrast masking, neighboring contrast is known to strongly effect human visual perception \cite{tariq}. 
This is why we are sometimes less likely to see a loss of resolution in highly textured areas as opposed to independent strong edges. 
Another very important aspect is that movement or temporal variation decreases our ability to detect and resolve contrast \cite{Ashraf2024castleCSFA}. 
This behavior is quantified by the dependency of the CSF on temporal frequencies as well as spatial frequencies.

\subsection{Visual Difference Predictors (VDP)}\label{sec:vdp}
Inspired by the compressive nature of the human vision, Visual Difference Predictors (VDP) are models that aim to predict the perceived differences between two images based on robustly modeling the frequency selective nature of the early visual system. The framework was originally introduced by Daly et al. \cite{Daly1992VisibleDP}. Since then, there have a lot of improvements and extensions to the original model. 
\cite{Mantiuk2011HDRVDP2AC} 
designed the HDR-VDP, which was one of the first human perception inspired metrics aim to quantify visual differences between HDR images. 
\cite{Mantiuk2021FovVideoVDP} then extended the VDP to account for human peripheral vision and color \cite{Mantiuk2024ColorVideoVDPAV}. 
\cite{Tursun2019LuminancecontrastawareFR} employed the VDP framework to control spatial quality in VR-HMDs. 
\cite{tariq} designed a variation of the VDP framework for real-time perceptually optimized tone mapping. Our HVPF can be thought of as a VDP framework specifically tailored for real-time application to the problem of Deep Learning based SR. To the best of our knowledge, our framework is the first application of robust Human Vision frameworks to efficient neural network based image/video processing.

%% file: sec/3_methods.tex
\section{Our Method: A Human Visual Processing Framework (HVPF)}\label{sec:methods}
Our approach is centered around the frequency domain interpretation of SR and the well-established fact that the early visual system is frequency-selective. The main motivation is that the difference between low and high-resolution images lies in the attenuation and removal of higher spatial frequencies. The task of an SR neural network is to reconstruct the missing high spatial frequencies. The better and stronger the neural network, the better the reconstruction. The main idea behind our work is that due to the limitations of the HVS, we do not always need a perfect reconstruction. If we can quantify the spectral nature of an SR method, we can guide the method using models of human visual perception to deliver the least expensive reconstruction required for optimal visual quality, i.e., any further improvement in reconstruction leads to no perceivable benefit while wasting computational resources.

While there are various ways to control the trade-off between the computational efficiency of an SR method and the reconstruction quality, we consider two: network branching and altering network depth. The first approach adds earlier exit points to the original network. Using earlier exit points, i.e.,  shallower branches, leads to less computation and lower reconstruction quality. In the latter method, different variants of an SR method are created by varying the depth of the original network to make shallower networks more efficient yet potentially comprising the visual quality of the output. Our method is not limited to the above techniques, and others, such as network quantization, could be easily incorporated. 

Given different variants of an SR method, our method aims to predict which version should be used in a specific region of an image or video frame. We propose to use attenuation curves to first quantify the reconstruction capability of a given variant. The attenuation curves express the ratio of the radially averaged 2D Fourier transform of the reconstructed output and its full-resolution counterpart. We demonstrated that such curves can be computed on a set of images and reused. Furthermore, we design a framework that expresses the required reconstruction quality in the form of the attenuation curve. Later, our method selects an appropriate SR variant to ensure adequate reconstruction quality for a region while minimizing computational costs. In practice, our method works on image patches which are both input to the SR method and our prediction. Below, we describe all the components of the method.

\subsection{Attenuation Response Estimation}
\label{sec:attenuation}
For a given SR method and an input image, we can quantify the quality of reconstruction by comparing the magnitude of the Fourier Transform of the reconstructed image and its ground-truth version. Given a ground-truth image $I$, we first downscale it by a factor $k$, producing image $I{\downarrow_k}$. 
Later, we use a SR method to upscale that image back to its original resolution. 
Comparing the Fourier transform of the resulting image $\hat{I} $ and the ground-truth counterpart $I$, allows us to quantify the reconstruction power of the analyzed SR method. 
More formally, given a SR method $\phi$ and a test image $I$, we define the frequency dependent attenuation curve as:
\begin{equation}
    \alpha^\phi_k(I,f) = \frac{|\mathscr{F}\left(\phi(I{\downarrow_k})\right)(f)|}{|\mathscr{F}\left( I\right)(f)|},
\end{equation}
where $\mathscr{F}$ denotes the Fourier transform.  Since we are interested in characterizing the reconstruction capability of the method, we do not compute the curve for one image but for a set of images $\{I_k\}$ and compute for a given SR method $\phi$ and downscaling factor $k$ the aggregated attenuation response curve as an average across all the images:
\begin{equation}
    \alpha^\phi_k(f) = \sum_{i=1}^N \alpha_k^\phi(I_i,f).
\end{equation}
Although not guaranteed, the value of $\alpha_k^\phi$ is expected to lie on $(0,1)$ range, where $\alpha_k^\phi(f) = 0$ means that the SR method was not able to reconstruct content at spatial frequency $f$, while $\alpha_k^\phi(f) = 1$ indicates the full capability in reconstructing this part of the signal. It has to be noted that this measure quantifies the presence of the signal in the reconstructed output and not its correctness. Nevertheless, we use this as a proxy for the reconstruction quality. This choice was further motivated by the fact that, although neural networks are non-linear functions, SR is a low-level task with consistent and deterministic characteristics in the frequency domain. The average spectrum of natural images is known to adhere to consistent Fourier characteristics like the inverse power law fall-off. Thus, an average attenuation curve is a good approximation for network response to a general natural image for the task of SR. In order to be more conservative, there is a possibility to use a particular percentile or quartile around the average, but our empirical analysis did not reveal it necessary, which is later confirmed by the results of our user experiments. 

\begin{figure}
    \centering
    \includegraphics[width=\linewidth]{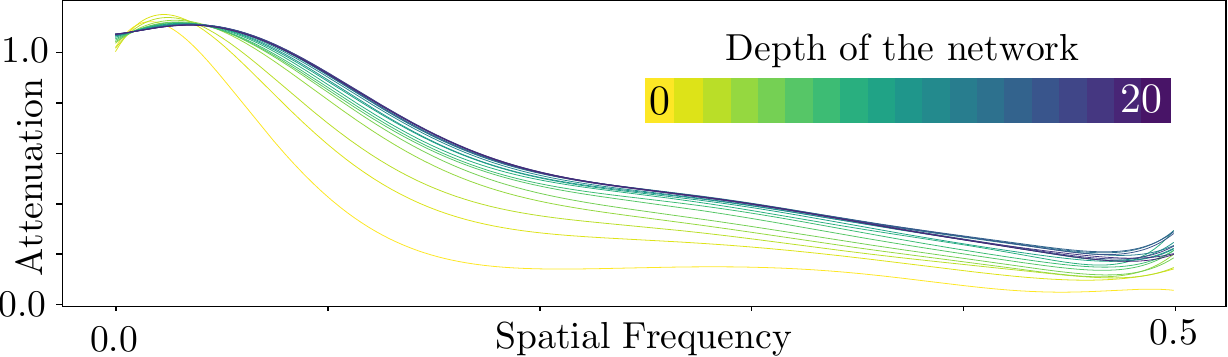}
    \caption{The attenuation curves derived for the case of modulating the depth (number of layers) of the network. 
    Depth equal 0 corresponds to a bicubic upsampling. As the depth of the network is reduced the performance of the solution in reconstructing high-spatial-frequency signal is reduced.}
    \label{figure:att_curves}
\end{figure}

For all the variants of the SR techniques considered in this work, we pre-compute the attenuation response curves using the above procedure. We always use 19 natural images from the set5~\cite{Bevilacqua2012LowComplexitySS} and set14~\cite{10.1007/978-3-642-27413-8_47} datasets, and compute different sets of curves for downscaling parameter $k\in\{2,4,8\}$. \cref{figure:att_curves} presents a set of curves for the case of varying the network depth. To obtain a more compact representations of the attenution functions, we model them using Gaussian fall-off: 
\begin{equation}
    \alpha'(f)=\frac{1}{a \sqrt{2 \pi}}\exp\left(-\frac{(f - b) ^ 2}{2a^2}\right) + c,
    \label{eq:1}
\end{equation}
where $f$ is the spatial frequency and $a,b,c$ are parameters estimated via fitting. After estimating the attenuation curves for variants of SR methods, the next step is to estimate which of them is ideal for a given patch of the image or video frame.

\subsection{Perceived Contrast Modeling}
\changes{We model the perceived luminance contrast following the VDP framework e.g. \cite{Tursun2019LuminancecontrastawareFR}. Please refer to the original work for more details on perceptual difference modeling. To summarize our model designed for application to SR, luminance contrast is computed as $C(f,p)$ using a multi-scale Laplacian-Gaussian pyramid, where $p$ is the location and $f$ is spatial frequency~\cite{Peli1990-vl,1095851}. The contrast measure is then normalized by the contrast sensitivity function (CSF), yielding $C_n(f,p)$. Finally, the perceived contrast, $C_t(f,p)$, also incorporates the visual masking model~\cite{Zeng2000PointwiseEV} with parameters $\alpha=0.7$ and $\beta=0.2$.}

\subsection{Optimization}
Given an input patch, 
the goal is to find the maximum attenuation that is under the resolution capabilities of human vision.
According to the contrast model, the attenuation remains undetectable by an observer if the contrast difference between the original image and the attenuated one is under 1 JND. Consequently, to find the attenuation that results in maximum performance gains yet imperceptible quality loss, we optimize the attenuation curve such that it results in exactly 1\,JND difference, i.e., 
\begin{equation}
\forall_f \,\, C_t'(p,f)-C_t(p,f) = 1,
\label{eq:2}
\end{equation}
where $C_t(p,f)$ represents the perceived contrast of the input image patch and $C_t'(p,f)$ represents the perceived contrast of the network output. Assuming the attenuation curves are a response of the network/branch to the input, the attenuation is a modulation of the physical image contrast at different frequencies:
\begin{equation}
    \alpha'(f)=\frac{C'(f,p)}{C(f,p)}=\frac{C'_n(f,p)}{C_n(f,p).}
\end{equation}
Our immediate goal is to estimate the tolerable output contrast $C'_n(f,p)$ from the constraint in~\cref{eq:2}.
We start by developing~\cref{eq:2} by substituting expressions from~\cite{Tursun2019LuminancecontrastawareFR}, which leads to the following form of the constraint:
\begin{equation}
    \resizebox{0.88\hsize}{!}{
    $\frac{\text{sign}\left(C'_n(f,p)\right)\cdot \left|C'_n(f,p)\right|^\alpha}{1+\frac{1}{|N|}\displaystyle\sum_{q\in N(p)}\left|C'_n(f,q)\right|^\beta}-\frac{\text{sign}\left(C_n(f,p)\right)\cdot \left|C_n(f,p)\right|^\alpha}{1+\frac{1}{|N|}\displaystyle\sum_{q\in N(p)}\left|C_n(f,q)\right|^\beta}=1$,}
    \label{eq:3}
\end{equation}
where the numerators encode the CSF-weighted contrast values, while the denominators model visual masking effect.
It is evident that $C_n'(f,q)$ cannot be directly calculated from this equation due to the visual masking term present in the denominator. To address this problem, similarly to the work of~\cite{Tursun2019LuminancecontrastawareFR}, we make the assumption that the contrast masking for the up-sampled output patch can be approximated by that of the input patch. Furthermore, knowing that the sign of the contrast should remain unaltered throughout the neural network processing~\cref{eq:3} we can derive $C_n'(f,p)$ directly as:
\begin{equation}
     \resizebox{0.88\hsize}{!}{
    $C_n'(f,p) = \left|\left(1+\sum_{q\in N(p)}\frac{\left|C_n(f,q)\right|^\beta}{|N|}\right)-\left|C_n(f,p)\right|^\alpha\right|^{1/\alpha}$}
    \label{eq:new_contrast}
\end{equation}
It should be noted that the sign of the contrast is omitted since the focus is on the magnitude of the contrast itself.
Now assuming that we have three levels of the contrast pyramid, tolerable attenuation at three different spatial frequencies can be calculated as follows:
\begin{equation}
t_i = \frac{C_n'(f_i,p)}{C_n(f_i,p)}, i\in{1,2,3}
\end{equation}
It is important to note that thanks to the derivation in \cref{eq:new_contrast}, the  attenuation can be computed directly from $C_n$, i.e., the input patch only.

Having the tolerable attenuation $t=\{t_1,t_2,t_3\}$, we can find the most suitable SR network/branch by identifying the one with most similar attenuation curve (Section \ref{sec:attenuation}, \cref{eq:1}). More formally, this step can be defined using following optimization problem:
\begin{equation}
    \text{branch/network} = \arg\max_{j}\left\{\frac{t\cdot \hat t_j}{\lVert t\rVert\lVert\hat t_j\rVert}\right\},
\end{equation}
where $\hat t_j = \{\alpha'(f_1), \alpha'(f_2), \alpha'(f_3)\}$ is the vector of the estimated attenuation produced by \cref{eq:1} for a given netwrok/brach $j$. 

\subsection{Model Efficiency}
In our application, it is essential that the overhead of the HVPF is minimal compared to the performance improvements provided by the SR method. Although we prototyped our solution using Python and evaluated it on a single CPU thread, it has been shown that a similar processing pipeline can be implemented efficiently on a GPU. This is primarily due to the nature of the computation, which consists of independent per-pixel operations (see~\cref{eq:new_contrast}). Additionally, the construction of the contrast pyramid can be accelerated using MIPMAP functionality. Rencently, 
~\cite{tariq} have demonstrated a HVPF on the Meta Quest 2 VR headset, achieving 2K resolution with a single execution time of under 1\,ms.

%% file: sec/4_results.tex
\section{Experimental Setup}\label{sec:results}
\begin{figure}
    \centering
    \includegraphics[width=\linewidth]{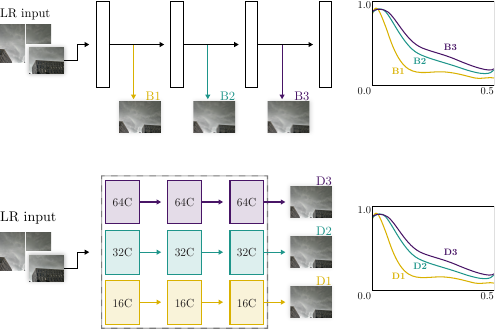} 
    \caption{Flowchart illustrating the methodology employed for efficient SR. The input image is divided into patches. For each patch, our task is to estimate which branch B (for the branching case) and which network D (for the depth case) should be used to each patch. The estimation is made through analyzing the computed attenuation characterstics (right). Such as to minimize computations with noticeable quality degradation.}
    \label{fig:flow-estimation}
\end{figure}
\begin{figure}
    \centering
    \includegraphics[width=\linewidth]{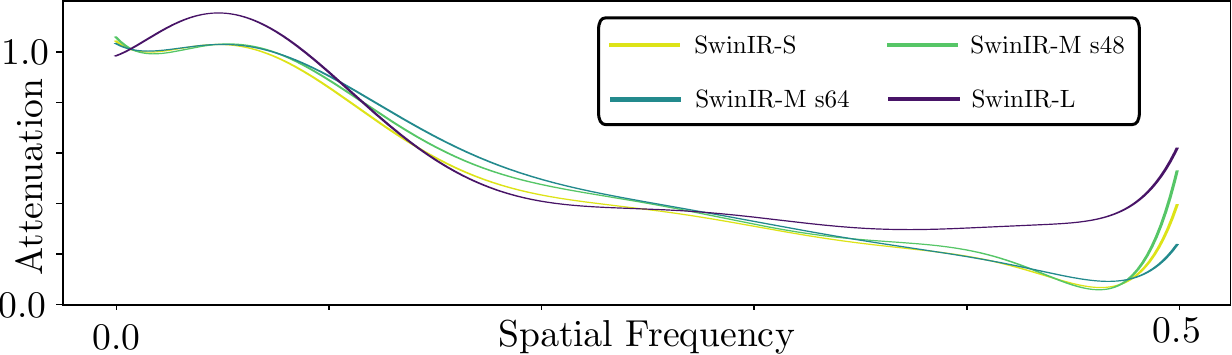}
    \caption{Attenuation curves derived from the SwinIR, a transformer based SR model. As the size of the model increases the capability of reconstructing higher frequency details increases.}
    \label{fig:transformer_attenuation}
\end{figure}

\paragraph{Validation on SR Models}
As mentioned in Section. 4, we employ our HVPF to optimize SR for two cases. The first approach adds 
earlier exit points (branches) to the original network. Using earlier  points, i.e., shallower branches, leads to less computation
and lower reconstruction quality. To test this case, we employ the popular and seminal VDSR~\cite{Kim2015AccurateIS} network. A neural network with 19 branched outputs was created as per the setting in~\cref{fig:flow-estimation} (top). After each ReLU activation function, an exit point was added, structured identically to the final layer of the original network., similar to~\cite{8851874}. Our task was to to use the HVPF to select the appropriate branch (per image patch) such that there is no noticeable quality loss.
\newline\indent The second approach is reducing the depth or number of channels of a network to make it more efficient. In this case, we have a number of candidate networks with different number of channels-per-layer, and our task is to employ our HVPF to select the appropriate one per image patch. To test this case, we employ the EDSR~\cite{lim2017enhanceddeepresidualnetworks} network. In the case of EDSR, five networks with varying numbers of channels per layer (256, 128, 64, 16, 8) were trained independently, the training procedure was the same as described in~\cite{lim2017enhanceddeepresidualnetworks}. Our task was to use the HVPF to select the appropriate network (per image patch) such that there is no noticeable quality loss. 
We include additional details on our choice of patch sizes in~\cref{sec:framework_input}.
\paragraph{On the generality of our visual framework.} We selected EDSR and VDSR models as test benchmarks for HVPF due to their widespread use, robustness and demonstrated efficacy over the years. Our algorithm is network agnostic: most SR approaches will follow similar attenuation characteristics, which are grounded in the nature of natural images and the Fourier nature of the SR problem~\cite{Mantiuk2024ColorVideoVDPAV}. 
In~\cref{fig:transformer_attenuation} we show attenuation curves for Transformer-based models, where increasing model size directly correlates with its ability to reconstruct higher spatial frequency content. 
This is also in line with previous research on implicit models for vision and graphics, where increasing the number of weights directly correlates with the capacity of the model to learn higher frequency content~\cite{10.1145/3503250} and together with our results on CNNs makes us confident on the generality of our HVPF to be leveraged with any SR method.

\paragraph{Subjective Quality Study.}
\label{sec:results_quality}
We developed a perceptual experiment with human subjects in order to validate our approach.
\newline\indent\emph{Task.} We employed a 2AFC experimental protocol. The task was a forced choice between two test cases in relation to a given high quality reference. Users were instructed to select the test case which was perceived more similar to the reference. On the right side we display the high quality reference; while the left side displayed either 
\begin{itemize}
    \item A) our HVPF-powered SR, selecting the appropriate network/branch for the corresponding image patch or,
    \item B) The output of the unaltered full deep network applied to the whole image.
\end{itemize}
These were displayed in randomized order. Users could use the space-bar to switch between A and B, and press ENTER when they had made their choice.
\newline\indent\emph{Stimuli.} We employed 24 high-quality natural scenes for our user study. 
The scenes were selected for diversity in characteristics such as luminance, contrast, and texture. 
\changes{Each scene was downscaled by a factor of $\times 4$ and upscaled back to the original resolution.}
\newline\indent In~\cref{sec:setup_quality} we included additional details on the experimental setup.
\section{Results and Discussion}
To evaluate the effectiveness of our method, we tested the model on images and videos, as our model is additionally capable of handling the temporal frequencies present in videos. 
\subsection{Quantitative Results}
The quantitative results for the image datasets are presented in~\cref{tab:comparisons_images}. 
The proposed method allows for comparable performance to the original networks while reducing the computational cost. 
For instance, the $\times$2 and $\times$4 upsampling operations exhibit a reduced computational cost ranging from 58\% to 22\% and from 70\% to 20\%, respectively.
\changes{For a single patch of size $10\times10$ the computational cost of our method is 39KFlops, while for a patch size of $35\times35$ is 477KFlops.}

Greater savings were achieved with $\times$2 upsampling in comparison to $\times$4 upsampling, which is in line with expectations. This is due to the increased presence of high-frequency information in the images, which allows for reconstructing certain parts of the image with less computational power. 

In order to evaluate the video, we estimated the optical flow. Subsequently, we calculated the temporal frequency that was necessary for our model, based on the velocities obtained with the optical flow. The frame rate considered for each video was 24 fps.
The results for the video datasets are presented in~\cref{tab:comparison_videos}. 
\newline\indent In~\cref{sec:flops} we provide an explanation for the choice of FLOPS as a measure of efficiency.
\begin{table*}
\caption{{\textbf{Quantitative comparison on image datasets}}}
  \centering
  \setlength\tabcolsep{3pt}
  \resizebox{\textwidth}{!}{
  \begin{tabular}{l c ccc ccc ccc ccc ccc}
    \toprule
        \multirow{2}{*}{Method} & \multirow{2}{*}{Scale}  & \multicolumn{3}{c}{Set5} & \multicolumn{3}{c}{Set14} & \multicolumn{3}{c}{BSD100} & \multicolumn{3}{c}{Urban100} & \multicolumn{3}{c}{DIV2K}\\
        \cmidrule(l{2pt}r{2pt}){3-5} \cmidrule(l{2pt}r{2pt}){6-8} \cmidrule(l{2pt}r{2pt}){9-11} \cmidrule(l{2pt}r{2pt}){12-14} \cmidrule(l{2pt}r{2pt}){15-17}
    & & PSNR$\uparrow$ & SSIM$\uparrow$& FLOPS$\downarrow$ & PSNR$\uparrow$ & SSIM$\uparrow$ & FLOPS$\downarrow$ & PSNR$\uparrow$ & SSIM$\uparrow$& FLOPS$\downarrow$  & PSNR$\uparrow$ & SSIM$\uparrow$ & FLOPS$\downarrow$ & PSNR$\uparrow$ & SSIM$\uparrow$& FLOPS$\downarrow$ \\
    \midrule
     Bicubic & $\times2$ & 32.32 & 0.923 &         & 28.60 & 0.859 &         & 28.22 & 0.834 &         & 25.48 & 0.840 &         & 31.23 & 0.898 &        \\
     VDSR & $\times2$ & 34.15 & 0.946 &  89.60G (100\%) & 29.98 & 0.899 & 173.68G (100\%) & 29.68 & 0.885 & 131.04G (100\%) & 27.17 & 0.900 & 144.99G (100\%) & 32.65 & 0.931 &  2.04T (100\%) \\
     VDSR-ours & $\times2$ & 32.53 & 0.936 &  \textbf{51.99G (58\%)} & 29.29 & 0.895 & \textbf{100.18G (57\%)} & 29.40 & 0.883 &  \textbf{61.63G (47\%)} & 26.48 & 0.889 &  \textbf{85.43G (58\%)} & 32.08 & 0.928 &  \textbf{1.06T (51\%)} \\
     EDSR & $\times2$ & 36.42 & 0.954 &   1.54T (100\%) & 32.03 & 0.905 &   2.97T (100\%) & 30.62 & 0.888 &   2.25T (100\%) & 30.42 & 0.932 &   2.56T (100\%) & 34.84 & 0.939 & 31.64T (100\%) \\
     EDSR-ours & $\times2$ & 36.00 & 0.951 & \textbf{353.09G (22\%)} & 31.58 & 0.901 & \textbf{691.09G (23\%)} & 30.26 & 0.882 & \textbf{466.39G (20\%)} & 29.74 & 0.925 & \textbf{606.62G (23\%)} & 34.25 & 0.925 &  \textbf{7.27T (22\%)} \\
     \midrule
     Bicubic & $\times4$ & 26.98 & 0.790 &         & 24.28 & 0.676 &         & 24.54 & 0.638 &         & 21.89 & 0.642 &         & 26.80 & 0.756 &         \\
     VDSR & $\times4$ & 28.13 & 0.827 &  89.60G (100\%) & 25.01 & 0.709 & 173.68G (100\%) & 25.11 & 0.670 & 131.04G (100\%) & 22.75 & 0.698 & 572.38G (100\%) & 27.52 & 0.786 &   2.04T (100\%) \\
     VDSR-ours & $\times4$ & 27.77 & 0.823 &  \textbf{64.13G (71\%)} & 24.86 & 0.713 & \textbf{118.64G (68\%)} & 25.08 & 0.670 &  \textbf{69.27G (52\%)} & 22.65 & 0.699 & \textbf{387.01G (67\%)} & 27.44 & 0.788 &   \textbf{1.29T (63\%)} \\
     EDSR & $\times4$ & 30.60 & 0.878 & 579.49G (100\%) & 26.95 & 0.753 &   1.00T (100\%) & 26.01 & 0.706 & 695.39G (100\%) & 24.82 & 0.776 &   3.16T (100\%) & 29.05 & 0.822 &  10.32T (100\%) \\
     EDSR-ours & $\times4$ & 30.27 & 0.874 & \textbf{145.34G (25\%)} & 26.69 & 0.750 & \textbf{236.71G (23\%)} & 25.82 & 0.701 & \textbf{173.54G (24\%)} & 24.48 & 0.765 & \textbf{746.13G (23\%)} & 28.75 & 0.816 &   \textbf{2.46T (23\%)} \\
    \bottomrule
  \end{tabular}}
  \vspace{2mm}
  \label{tab:comparisons_images}
\end{table*}
\begin{table*}
\caption{{\textbf{Quantitative comparison on video datasets X4 upscaling}}}
  \centering
  \setlength\tabcolsep{3pt}
  \resizebox{\textwidth}{!}{
  \begin{tabular}{l ccc ccc ccc}
    \toprule
        \multirow{2}{*}{Method}  & \multicolumn{3}{c}{REDS} & \multicolumn{3}{c}{Vid4} & \multicolumn{3}{c}{UDM10} \\
        \cmidrule(l{2pt}r{2pt}){2-4} \cmidrule(l{2pt}r{2pt}){5-7} \cmidrule(l{2pt}r{2pt}){8-10}
    & PSNR$\uparrow$ & SSIM$\uparrow$& FLOPS$\downarrow$ & PSNR$\uparrow$ & SSIM$\uparrow$ & FLOPS$\downarrow$ & PSNR$\uparrow$ & SSIM$\uparrow$& FLOPS$\downarrow$ \\
    \midrule
     Bicubic & 26.39 & 0.724 &         & 22.44 & 0.614 &         & 30.76 & 0.884 &         \\
     
     VDSR & 27.11 & 0.756 & 702.24G (100\%) & 23.14 & 0.670 & 291.96G (100\%) & 31.71 & 0.899 & 680.96G (100\%) \\
     VDSR-ours w/o temporal frequency & 27.03 & 0.755 & 443.84G ( 63\%) & 23.06 & 0.667 & 194.23G ( 66\%) & 31.62 & 0.903 & 433.30G ( 63\%) \\
     VDSR-ours & 26.66 & 0.739 & \textbf{173.77G ( 24\%)} & 23.05 & 0.667 & \textbf{193.00G ( 66\%)} & 31.58 & 0.902 & \textbf{386.77G ( 56\%)} \\
     
     EDSR & 28.27 & 0.791 &   3.24T (100\%) & 23.92 & 0.711 &   1.59T (100\%) & 34.28 & 0.929 &   3.24T (100\%) \\
     EDSR-ours w/o temporal frequency & 27.03 & 0.755 & 813.12G ( 25\%) & 23.73 & 0.703 & \textbf{348.84G (21\%)} & 33.78 & 0.924 & 813.96G (25\%) \\
     EDSR-ours & 27.03 & 0.755 & \textbf{586.58G ( 18\%)} & 23.73 & 0.703 & \textbf{348.84G (21\%)} & 33.84 & 0.925 & \textbf{801.61G (24\%)} \\
    \bottomrule
  \end{tabular}}
  \vspace{2mm}
  \label{tab:comparison_videos}
\end{table*}
\begin{figure}
    \centering
    \includegraphics[width=\linewidth]{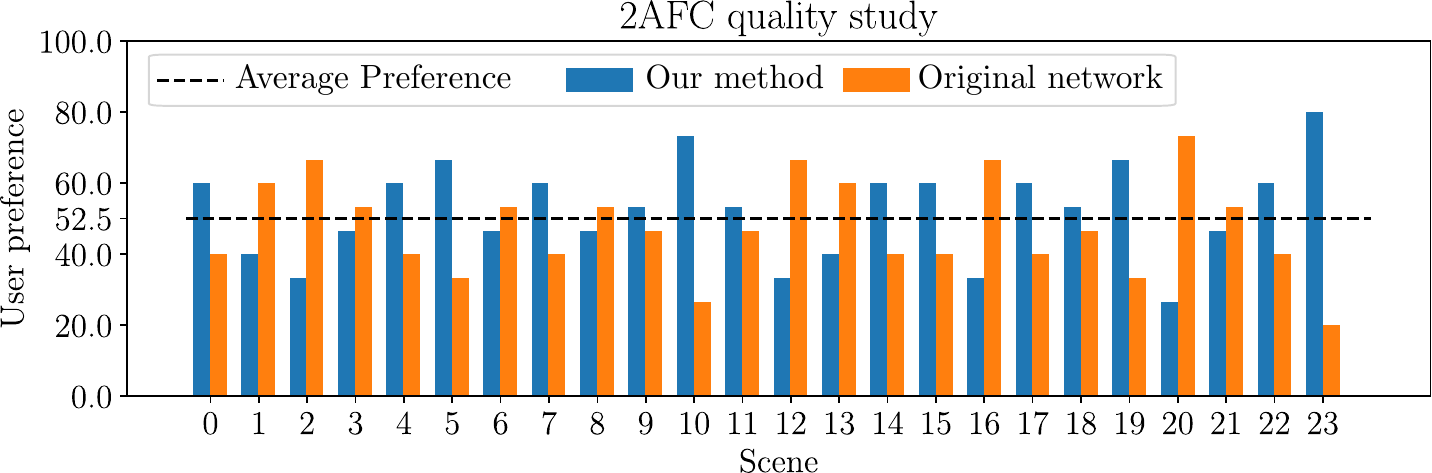}
    \caption{The result of our subjective study (for 15 participants) for the network branching application.}
    \label{fig:user-experiment}
\end{figure}
\subsection{Subjective Quality Results}
\begin{figure}
    \centering
    \includegraphics[width=\linewidth]{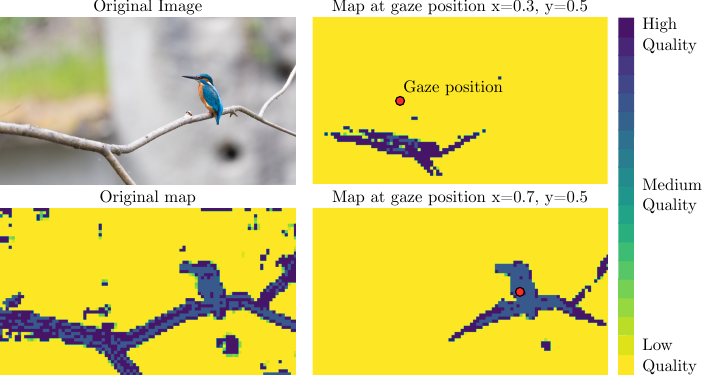}
    \caption{Our model predictions based on gaze position with $\times 4$ super-resolution. In the first column, we have the original image and the corresponding quality map. In the other column we have the quality maps at different gaze positions.}
    \label{fig:vr}
\end{figure}

It is well-known that metrics such as PSNR, SSIM, etc are not correlated with human quality assessments, unable to model the intricacies of how humans perceive image quality~\cite{10.1145/3528233.3530729,7351345,nilsson2020understanding}. As our HVPF is based on a robust and detailed modeling of early visual processing in a scene dependent manner (unlike heuristic metrics trained over a large set of images e.g~\cite{zhang2018unreasonable}), our hypothesis (little noticeable quality loss) can only be aptly verified using subjective quality studies.
\newline\indent\cref{fig:user-experiment} shows the results of our 2AFC user study on images for the network branching application case on the VDSR. It can be seen that on average, the preference value hovers around 50$\%$, which is indicative that users were not able to perceive any difference between the test cases A and B in relation to the reference, which supports our initial goal and hypothesis. As expected, no differences were perceived even with the reported differences in PSNR and SSIM in \cref{tab:comparisons_images}. %
Further results for the study on videos are provided in~\cref{sec:sub_results:video}, which demonstrates the efficacy of our HVPF for handling motion too.
\newline\indent A more detailed study on the network channel depth SR alternative (EDSR, bottom~\cref{fig:flow-estimation}) is presented in~\cref{sec:sub_results:imgs}, providing similar conclusions. In summary, the studies demonstrate that the results indeed conform with our hypothesis that there is no perceivable loss in visual quality, even when there are significant computational savings through the application of our HVPF. 
\newline\indent Finally, we also include promising early results on foveated SR (super resolution aware of gaze position, leveraging the substantially degraded visual acuity on the peripheral vision) in~\cref{fig:vr} and in~\cref{sec:ar}.

%% file: sec/4.5_future_work.tex
\section{Limitations and Future work}
Although the method demonstrated satisfactory performance with regard to video content, it was observed that aliasing issues were present. This phenomenon is common when individually processing frames as isolated still images.
In addition, our method does not explicitly guarantee the spatial consistency between the patches. However, no such inconsistencies were noted or reported by the participants in our experiments.
It would be of interest to investigate the potential of our HVPF in conjunction with video-specific SR techniques. 
Evaluating the performance of our model within foveation-aware frameworks is also an interesting avenue for future work, given that our choice of contrast-sensitivity function (StelaCSF) is capable of modelling contrast sensitivity based on eccentricity.
Finally, our estimation is based solely on image luminance; we could further leverage additional information, such as colour, by integrating recent advances in color-aware CSFs like CastleCSF~\cite{10.1167/jov.24.4.5}.

%% file: sec/5_conclusion.tex
\section{Conclusions}\label{sec:conclusion}
We present a thorough novel framework to leverage the specific deficiencies of the HVS to optimize computational resources in the context of SR. 
HVPF is fast, robust, and can be seamlessly integrated into any SR framework, adaptively optimizing computational resources to the areas in the image that require it, from the assumption that a human will be the final observer of the resulting SR image. 
While normally SR methods are evaluated in terms of reconstruction quality, through metrics such as PSNR or SSIM, these metrics do not model the intricacies of human visual quality assessment, and have been repeatedly demonstrated as uncorrelated with human visual perception~\cite{10.1145/3528233.3530729,7351345,nilsson2020understanding}. In contrast, we validate our framework through a series of human studies, showcasing indistinguishable quality at substantially reduced computational costs. Our results demonstrate that even with FLOP reductions by factors of-and-greater than 3x, our HVPF minimally degrades the SR output such that the final result is not visually distinguishable from the output of a fully-resourced deep network. 
We are confident that our method can be further extended in the future to better integrate with video-specific SR approaches, and will be particularly relevant on VR and AR eccentricity-aware SR frameworks, where computational savings can be pushed significantly further.

%% file: sec/X_suppl.tex
\renewcommand*{\thesection}{\Alph{section}}
\clearpage
\setcounter{page}{1}
\setcounter{section}{0}
\maketitlesupplementary
\begin{abstract}
This supplementary file presents further details and additional results of the proposed method. These results illustrate further outcomes of the user study. Subsequently, the preliminary prototype of the method applied to AR/VR headsets are illustrated. Finally, further qualitative results of the perceptual model are presented to demonstrate the efficacy of the method.
\end{abstract}

\section{Subjective Quality Study Setup}
\label{sec:setup_quality}
\indent\emph{Setup.} For our experiments, we assumed standard office conditions where the content is viewed on a 27-inch Dell U2723QE display with a resolution of 3840 $\times$ 2160 and a peak luminance of 400 $cd/m^2$, from a viewing distance of 60\,cm. All our results are calculated according to this setup. During our experiments with human subjects, the viewing distance was controlled with the use of a chin rest that allowed to maintain constant viewing conditions throughout the experiments for all participants.

\section{Additional Subjective Quality Results}
\label{sec:sub_results}
\subsection{Channel Depth Application}
\label{sec:sub_results:imgs}
\begin{figure}
    \centering
    \includegraphics[width=\linewidth]{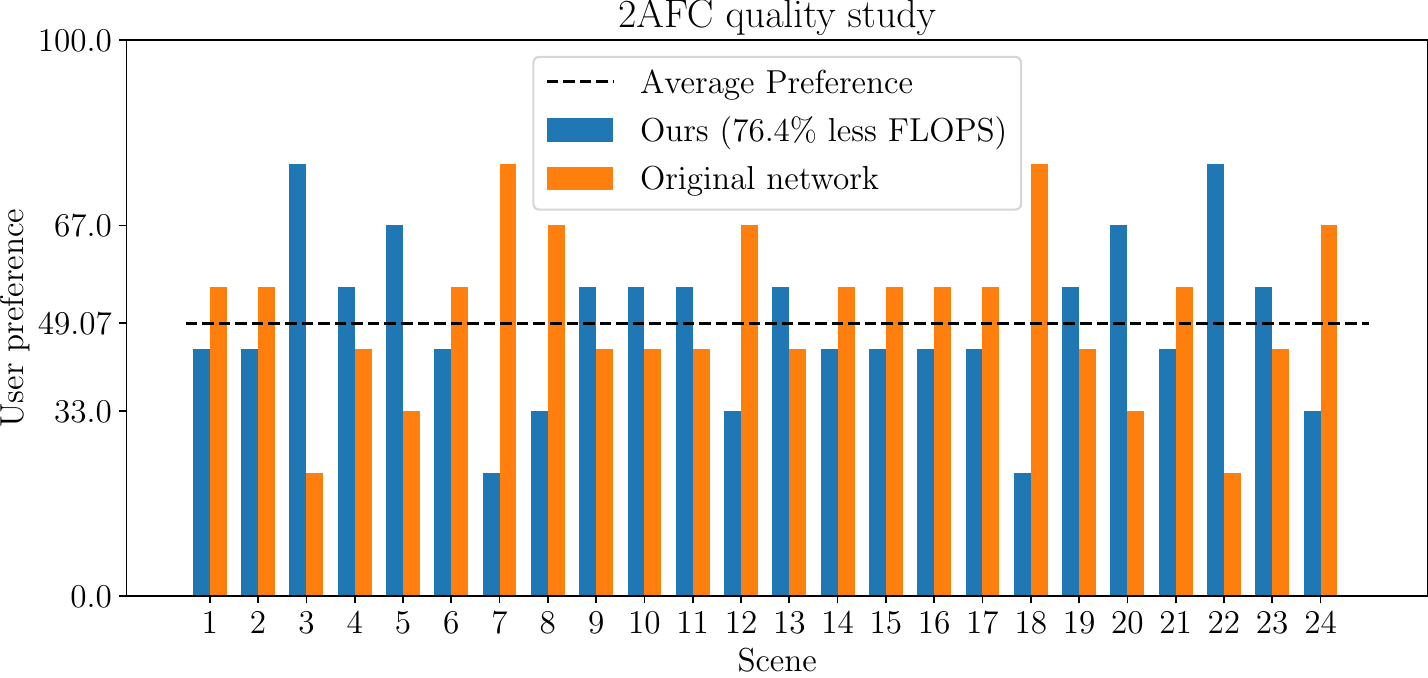}
    \caption{The result of our subjective study (for 9 participants)
for the network channel depth application.}
    \label{fig:experimet_edsr}
\end{figure}

For this experiment, we used the same scenes as described in \cref{sec:results_quality}. We up-scaled the images using a set of EDSR networks. We trained 5 different networks, with either (256, 128, 64, 16, 8) channels per-layer. The baseline was an EDSR with 256 channels per layer applied uniformly across the whole image. The baseline was compared with SR controlled using our perceptual model, which selected one of the 5 candidate networks per each patch. \cref{fig:experimet_edsr} shows the results of our 2AFC user study, it can be seen that on average, the preference value hovers around 50\%, which is indicative that users wone average not able to perceive any difference between the test cases A and B in relation to the reference. The indicates that our perceptual model controlled the results such that any quality loss is not visible, while using 76.4$\%$ less FLOPS. 

\subsection{Video Content}
\label{sec:sub_results:video}
\textbf{Setup and Task} The same experimental protocol as described in \cref{sec:results_quality} was employed for this experiment. 
\newline
\textbf{Stimuli} The stimuli were derived from seven natural videos from the Inter4k \cite{stergiou2022adapoolexponentialadaptivepooling} dataset. 
\newline
\textbf{Data preparation} Each video frame was down-sampled by a factor of eight and then up-sampled by a factor of four using the VDSR network, specifically the network branching application.

\begin{figure}
    \centering
    \includegraphics[width=\linewidth]{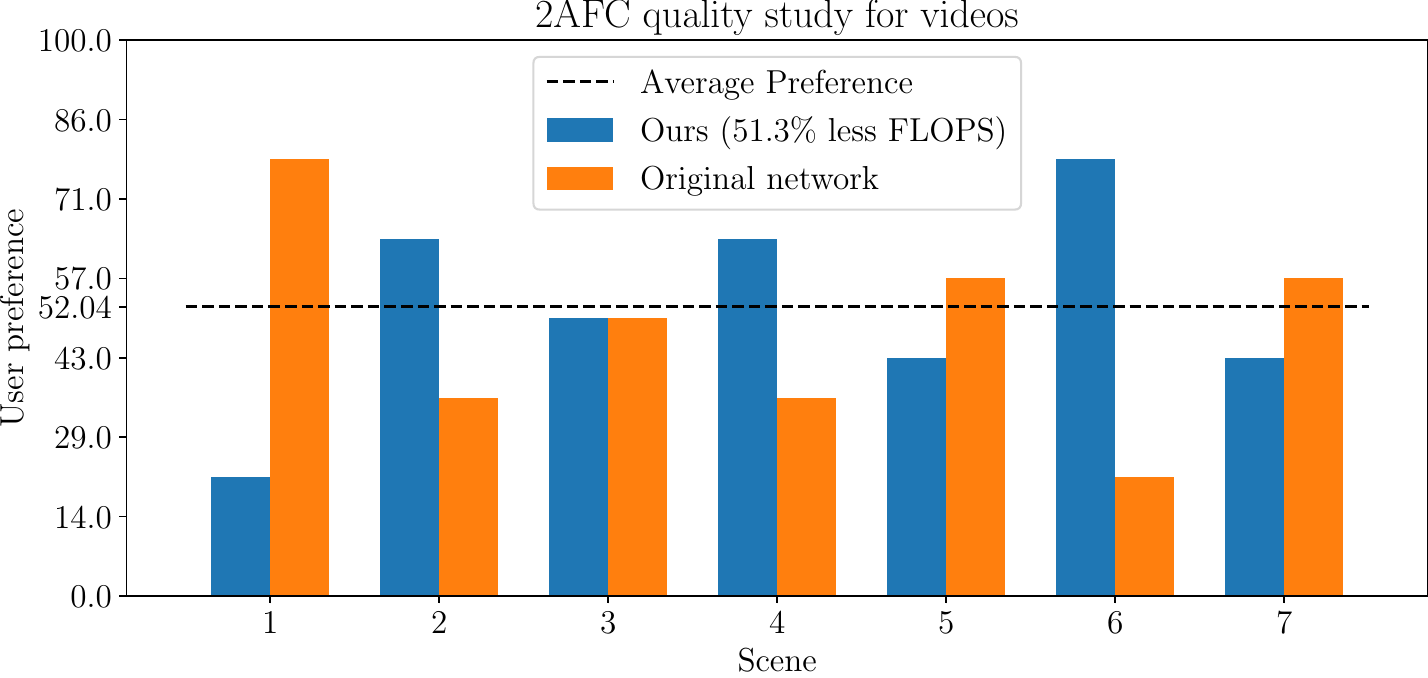}
    \caption{The result of our subjective study (for 14 participants)
for the network branching application for videos.}
    \label{fig:experimet_video_vdsr}
\end{figure}

\cref{fig:experimet_video_vdsr} shows the findings of our 2AFC user study. It can be observed that the mean preference is around 50\%, which suggests that users were unable to perceive a difference between test cases A and B in relation to the reference, even when our method uses 51.3$\%$ less FLOPS. 

\changes{\subsection{Information about participants}
All participants were recruited from CS department, aged 20 to 30.
The network branching user study was conducted with a total of 15 participants (4F, 11M).
The channel depth user study was conducted with a sample of 9 participants (3F, 6M)).
Finally, the network branching applied to videos user study was conducted with 14 participants (3F, 11M).
They had normal or corrected-to-normal vision and were unaware of the experiment's purpose.}

\changes{\section{FLOPS for evaluation and performance bottleneck}
\label{sec:flops}
We decided to use the average FLOPS for efficiency due to its universality, independence from specific machine characteristics and implementation, as well as the fact that this measure has been used in similar studies~\cite{kong2021classsr,jeong2024accelerating,xie2021learning,wang2022adaptive}. 
In our method patches that require the use of a network with larger size/capacity could be considered as a bottleneck, and in such a case, using the average FLOPS does not convey this information. A wall clock time elapsed could be considered a better measure of performance, for our method we would like to stress that the execution of the full network will not necessarily be the bottleneck. In scenarios where the number of processors is smaller than the number of patches, scheduling will take care of evenly distributing the load. 
A solution to further distribute the load more evenly would be to consider multiple consecutive frames for processing. Furthermore, it is possible that no patch in an image requires full network, and the full network will never be used.
}

\changes{\section{Framework input} 
\label{sec:framework_input}
The input patch size (size of patches into which the image is divided, as shown in \cref{fig:flow-estimation}) to our \changes{HVPF} is equivalent to the size of the receptive field of the upsampling model employed. In the case of the VDSR network, the input patch is $\frac{40}{k}\times\frac{40}{k}$ pixels. 
This is due to the fact that, during the \changes{HVPF} prediction, the low-resolution image is being considered. Before being conveyed to the VDSR network, the LR image is upsampled through bicubic interpolation.
Consequently, in the event of $\times$4 upsampling, the input to the \changes{HVPF} is 10x10 pixels, corresponding to a patch of 40x40 pixels in the image upsampled with bicubic interpolation. 
In the case of the EDSR network, the input patch is 48 $\times$ 48 pixels, taken from the low-resolution image. As no prior upsampling is involved in the input image, there is no need to consider a lower-size patch. In certain instances, utilizing input patches smaller than the neural network's receptive field may be advantageous, particularly in the context of smaller images.}

\section{Extension - AR/VR Display}
\label{sec:ar}
Next generation standalone virtual/augmented reality headsets demand high spatial quality, refresh-rate and power efficiency in real-time. Our framework can be applied for gaze-contingent super-resolution for AR/VR headsets. The main justification is that for wide field-of-view displays, human visual acuity decreases significantly away from the gaze-location (fovea). This inhomogeneity is frequently associated with the distribution of retinal cells across the visual field, as demonstrated by \cite{Curcio1990-ti,Watson2014-am}.

\begin{figure*}
    \centering
    \includegraphics[width=\linewidth]{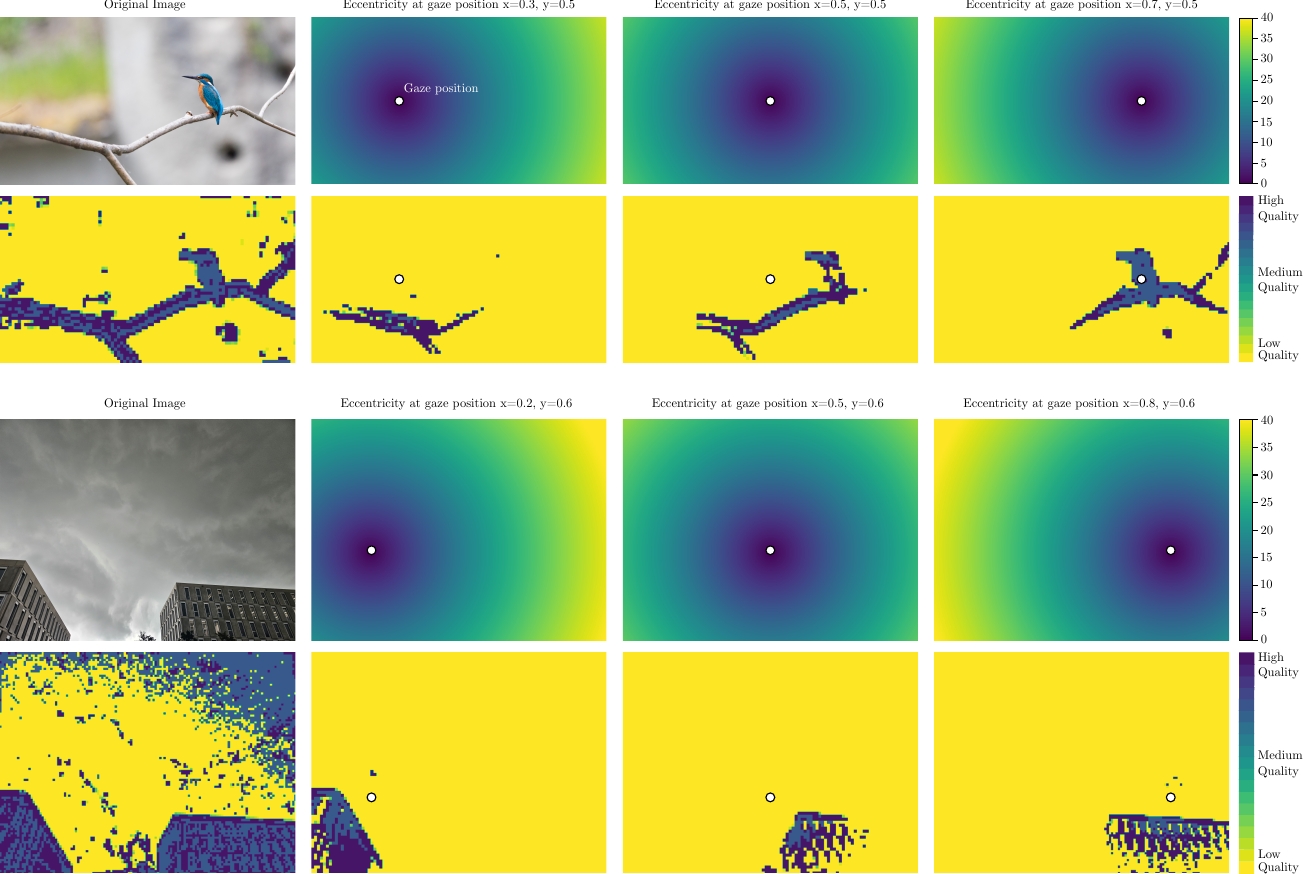}
    \caption{Our model predictions based on gaze position with $\times 4$ super-resolution. In the first column, we have the original image and the corresponding quality map. In the other columns we have on top the eccentricity map expressed in degrees and bottom we have the corresponding quality map.}
    \label{fig:maps_with_eccentricity}
\end{figure*}

Contrast sensitivity models such as the StelaCSF appropriately model human contrast perception as a function of eccentricity, and thus can extend our model to account for acuity across the visual field. Modern VR/AR headsets have built in eye-trackers that can be used to control our framework. 
In \cref{fig:maps_with_eccentricity}, we present some preliminary results for our quality map estimation with different gaze positions on the screen. The top row shows the eccentricity map relative to the gaze location, and the bottom row shows how our prediction for required SR quality varies. As anticipated, our perceptual model predicts that higher quality resolution will be used when the user is looking, while for areas in the periphery, our model predicts that the lowest up-sampling quality will be used. The main application is rendering in a lower resolution throughout the field of view, and then up-sampling the rendering for real-time VR/AR displays using our technique. 

\section{More Qualitative Results}
\label{sec:more_results}
\begin{figure*}
    \centering
    \includegraphics[width=\linewidth,height=\textheight,keepaspectratio]{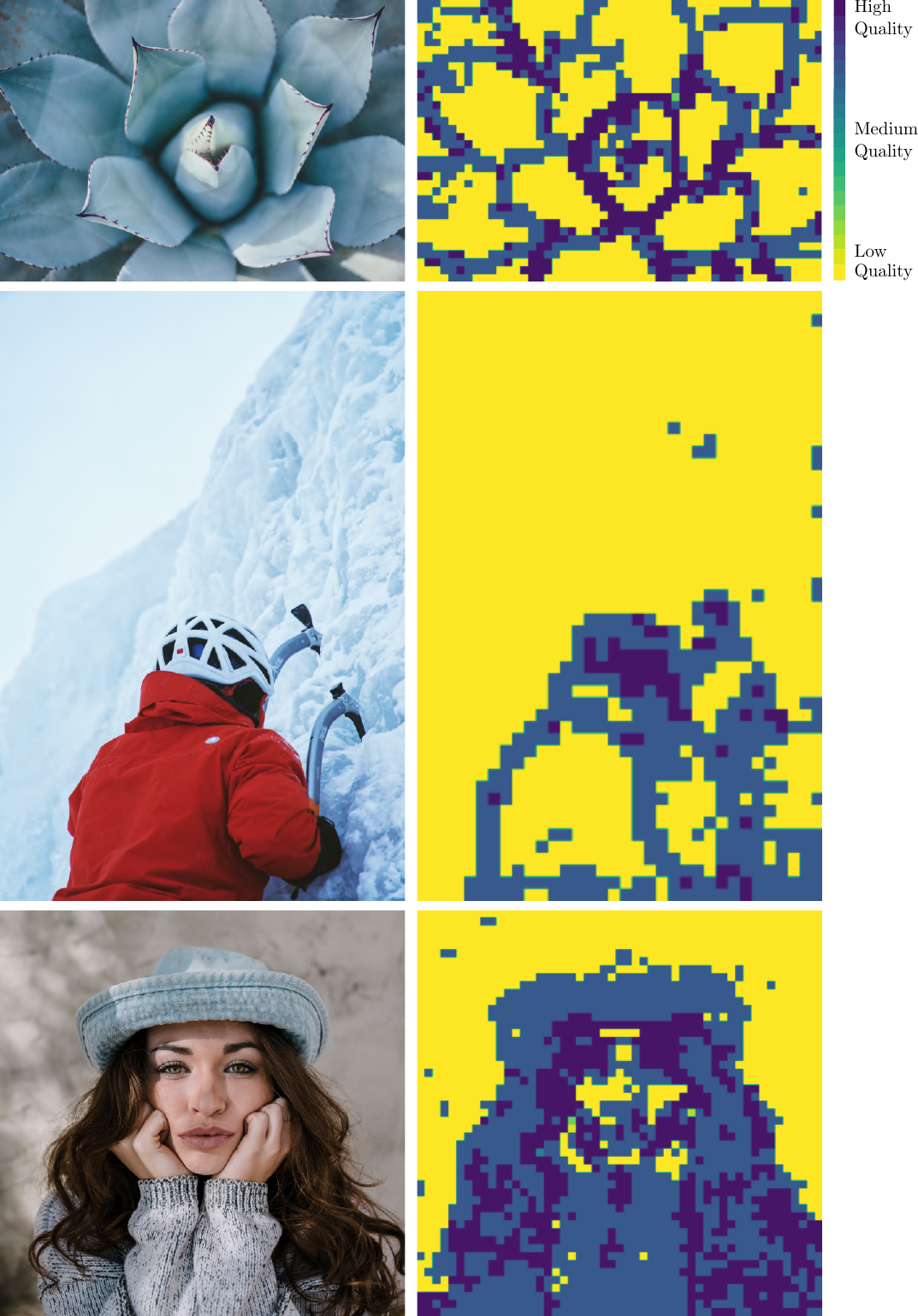}
    \caption{Our method quality map prediction results for $\times 4$ upsampling.}
    \label{fig:more_maps}
\end{figure*}

\begin{figure*}
    \centering
    \includegraphics[width=\linewidth,height=\textheight,keepaspectratio]{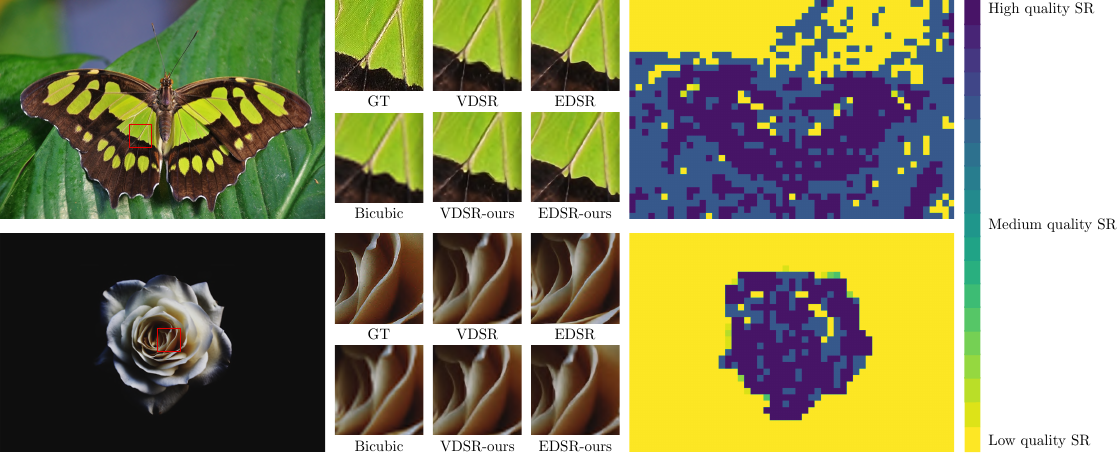}
    \caption{Visual results of our method compared to the original networks. On the right, we can observe the maps produced by our perceptual model.}
    \label{fig:more_images}
\end{figure*}

\begin{figure*}
    \centering
    \begin{subfigure}[b]{0.49\textwidth}
     \centering
     \includegraphics[width=\textwidth]{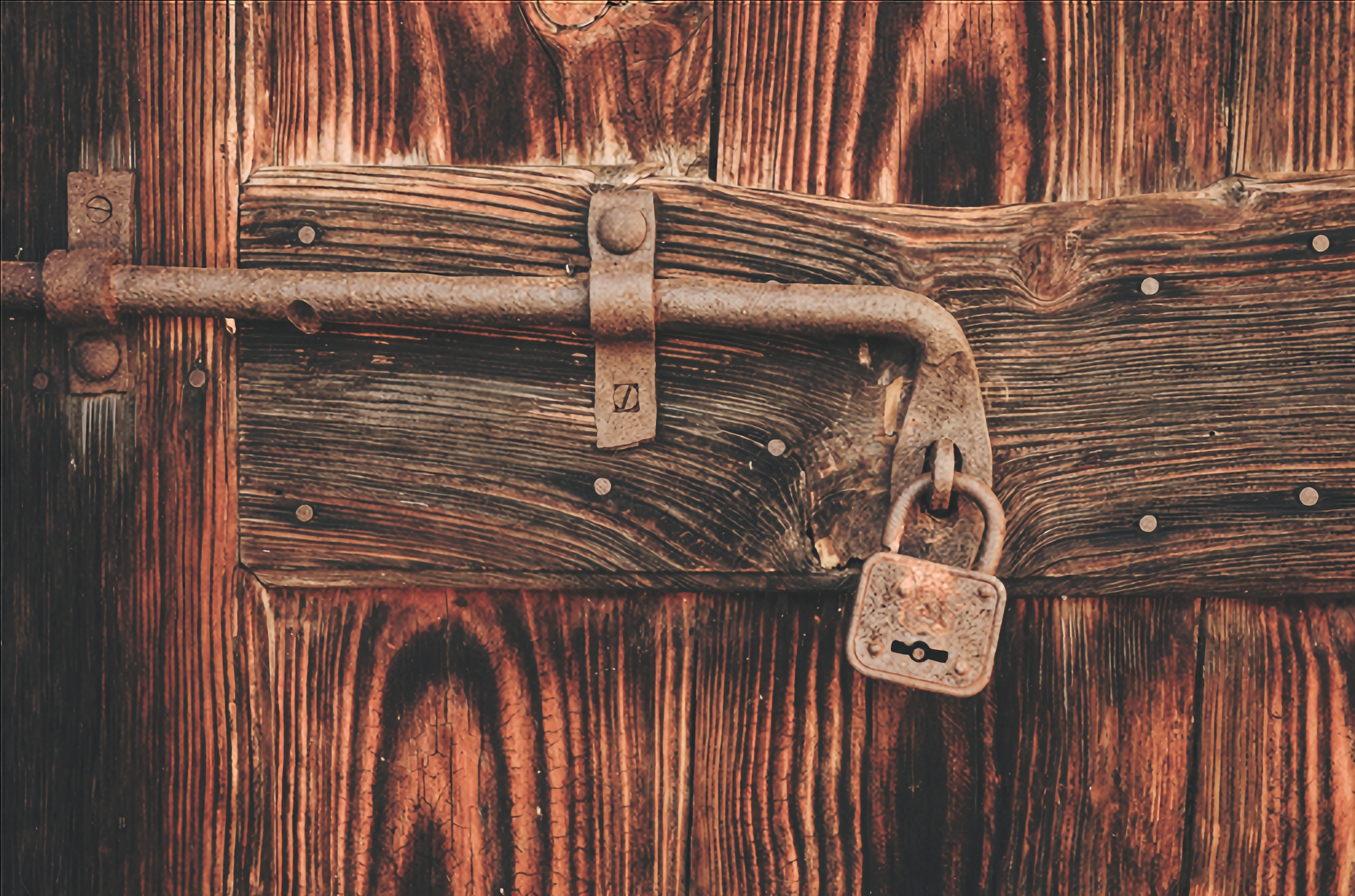}
     \caption{Original Network}
     \label{fig:o0885}
    \end{subfigure}
    \hfill
    \begin{subfigure}[b]{0.49\textwidth}
     \centering
     \includegraphics[width=\textwidth]{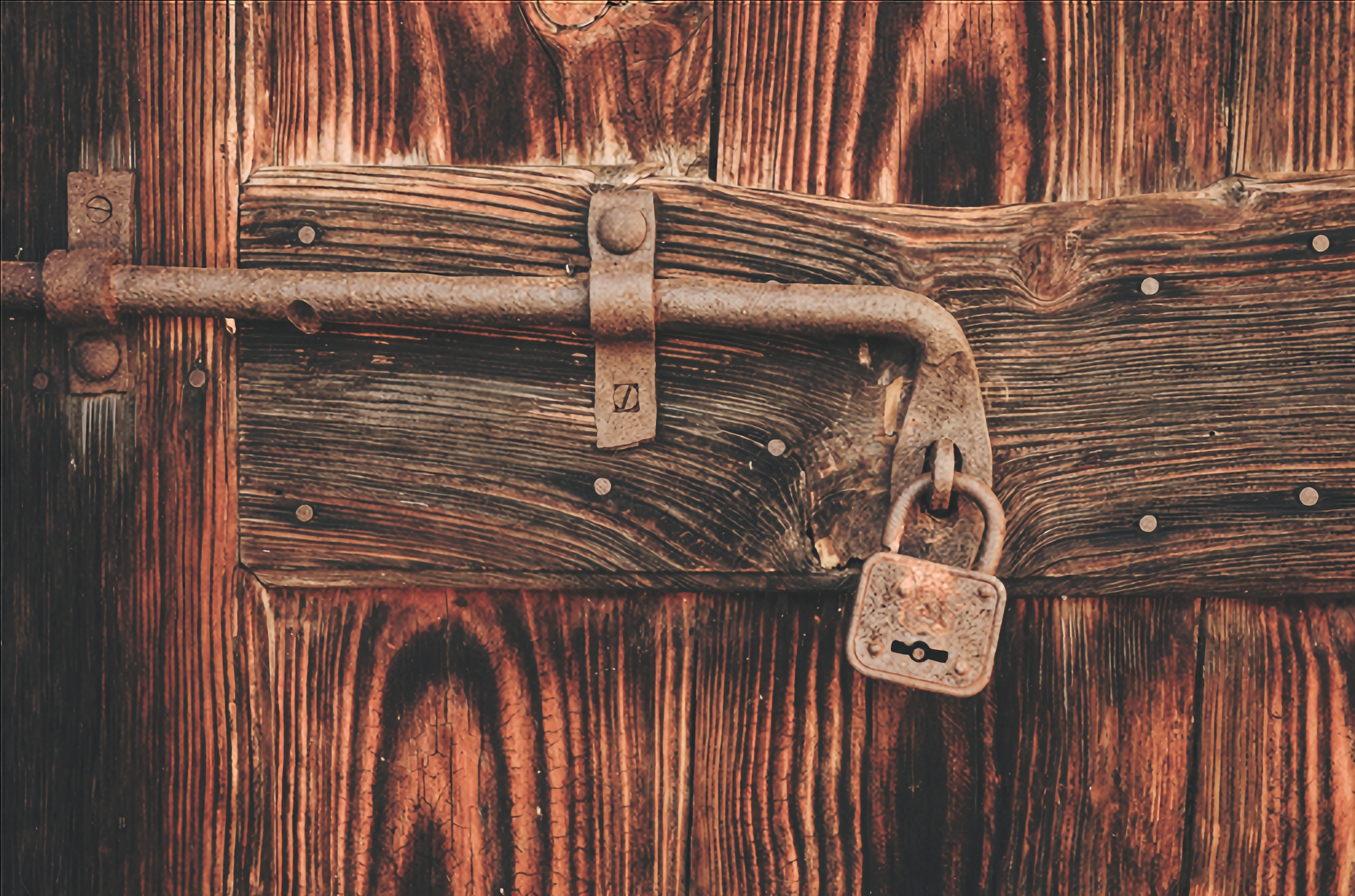}
     \caption{Our method}
     \label{fig:our0885}
    \end{subfigure}
    \caption{The SR result ($\times$4) of VDSR network and our method applied to VDSR network for image DIV2K-0885.}
    \label{fig:0885}
\end{figure*}

\begin{figure*}
    \centering
    \begin{subfigure}[b]{0.49\textwidth}
     \centering
     \includegraphics[width=\textwidth]{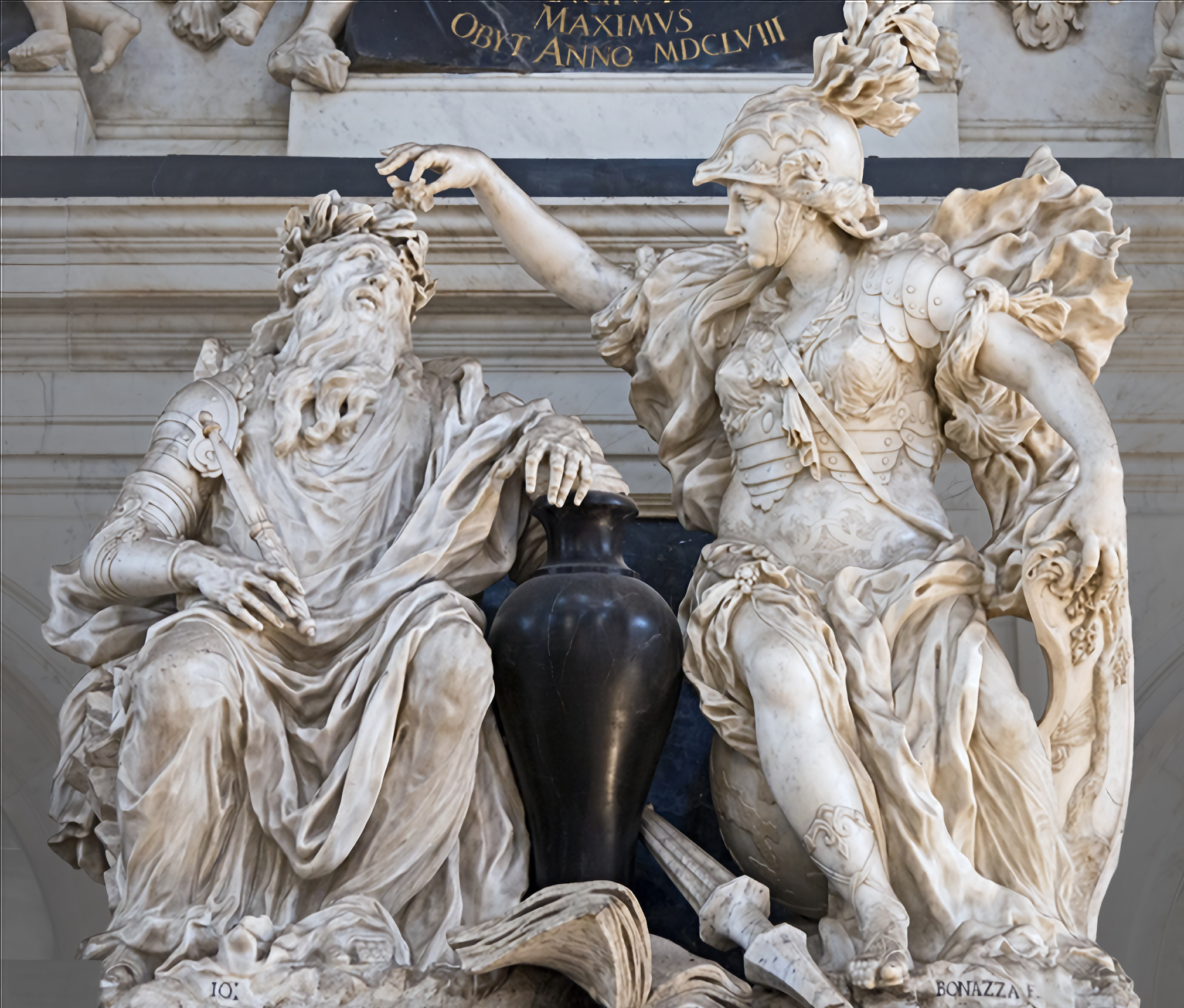}
     \caption{Original Network}
    \end{subfigure}
    \hfill
    \begin{subfigure}[b]{0.49\textwidth}
     \centering
     \includegraphics[width=\textwidth]{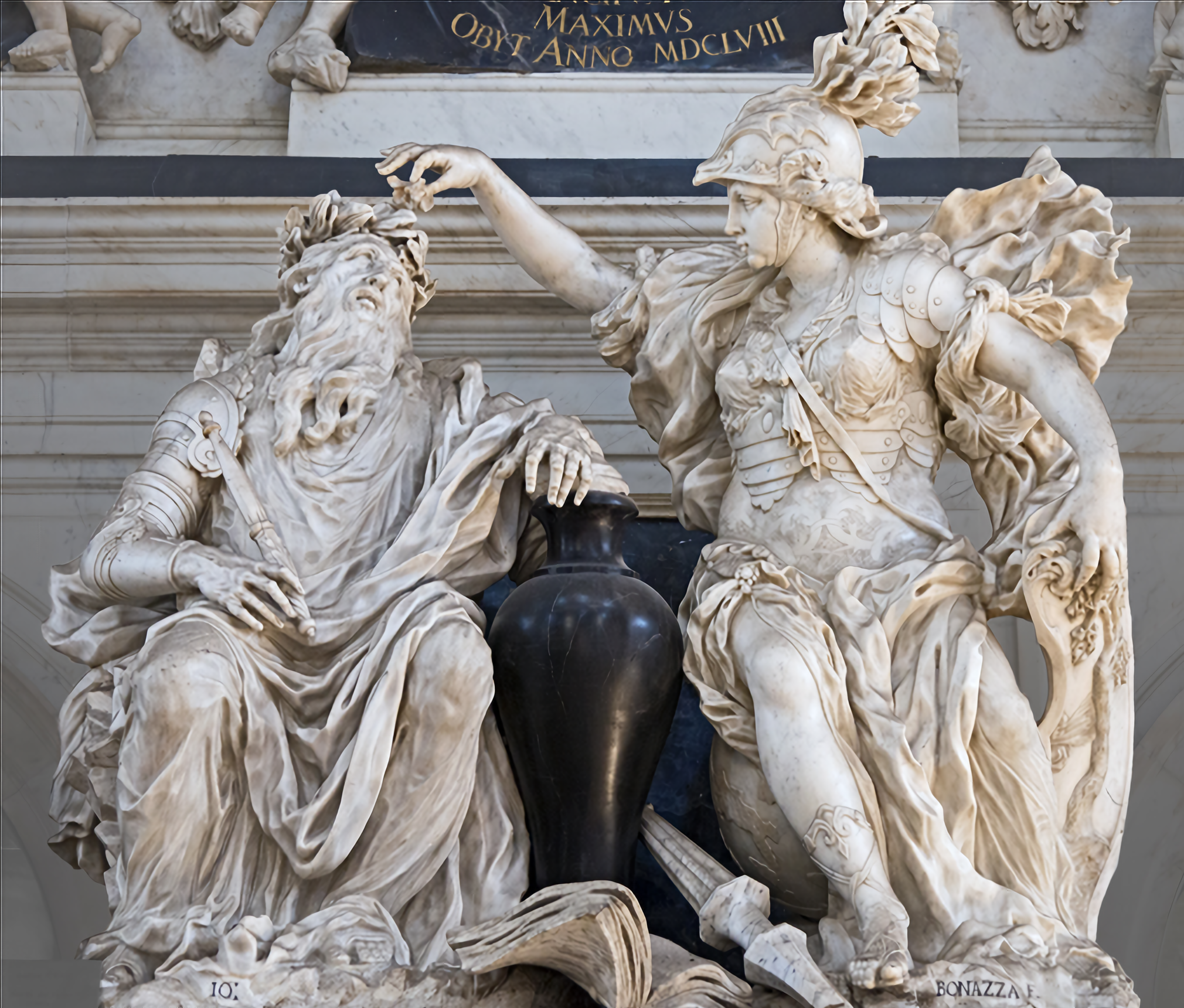}
     \caption{Our method}
    \end{subfigure}
    \caption{The SR result ($\times$4) of VDSR network and our method applied to VDSR network for image DIV2K-0878.}
    \label{fig:0878}
\end{figure*}

\begin{figure*}
    \centering
    \begin{subfigure}[b]{0.49\textwidth}
     \centering
     \includegraphics[width=\textwidth]{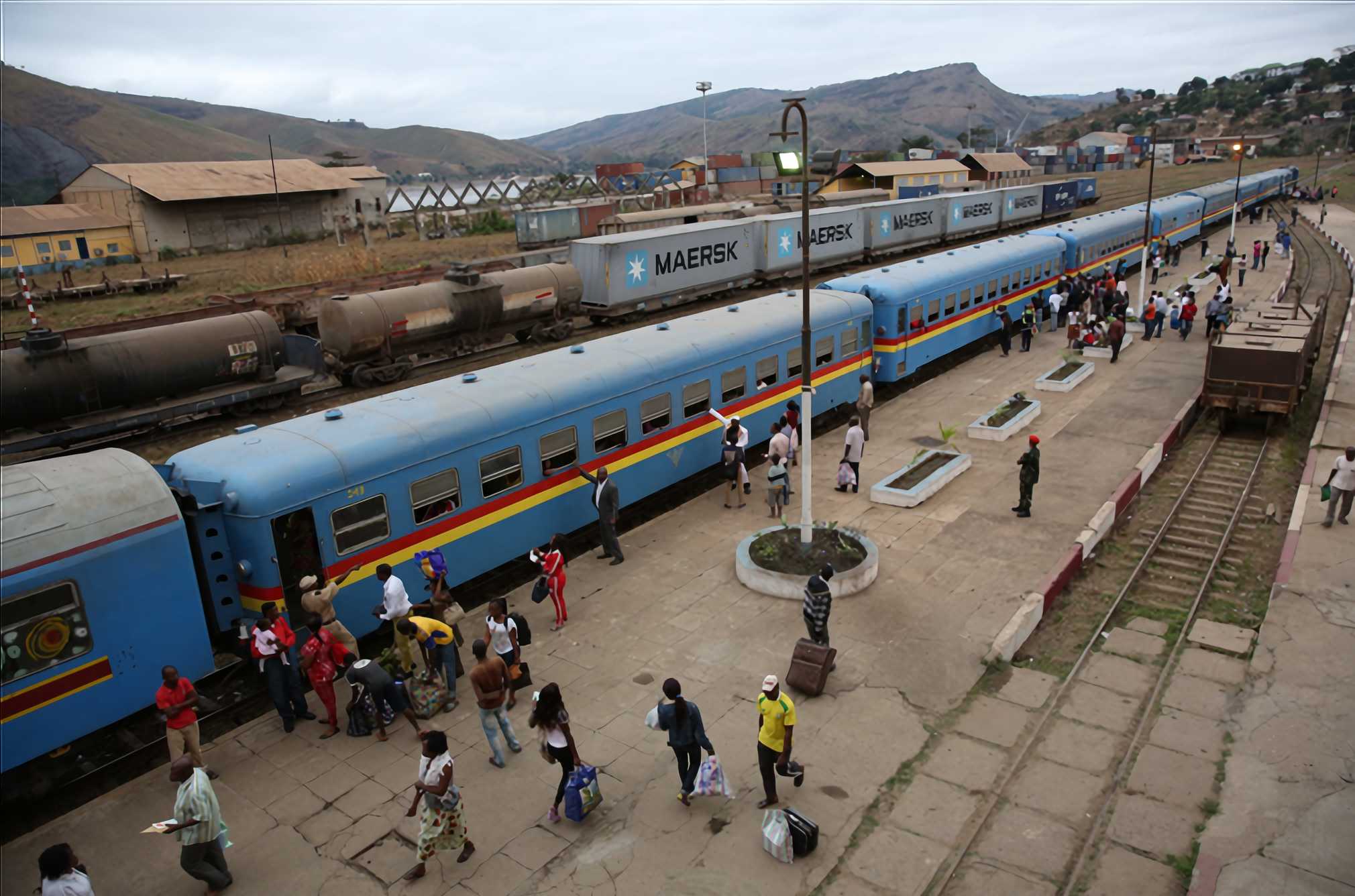}
     \caption{Original Network}
    \end{subfigure}
    \hfill
    \begin{subfigure}[b]{0.49\textwidth}
     \centering
     \includegraphics[width=\textwidth]{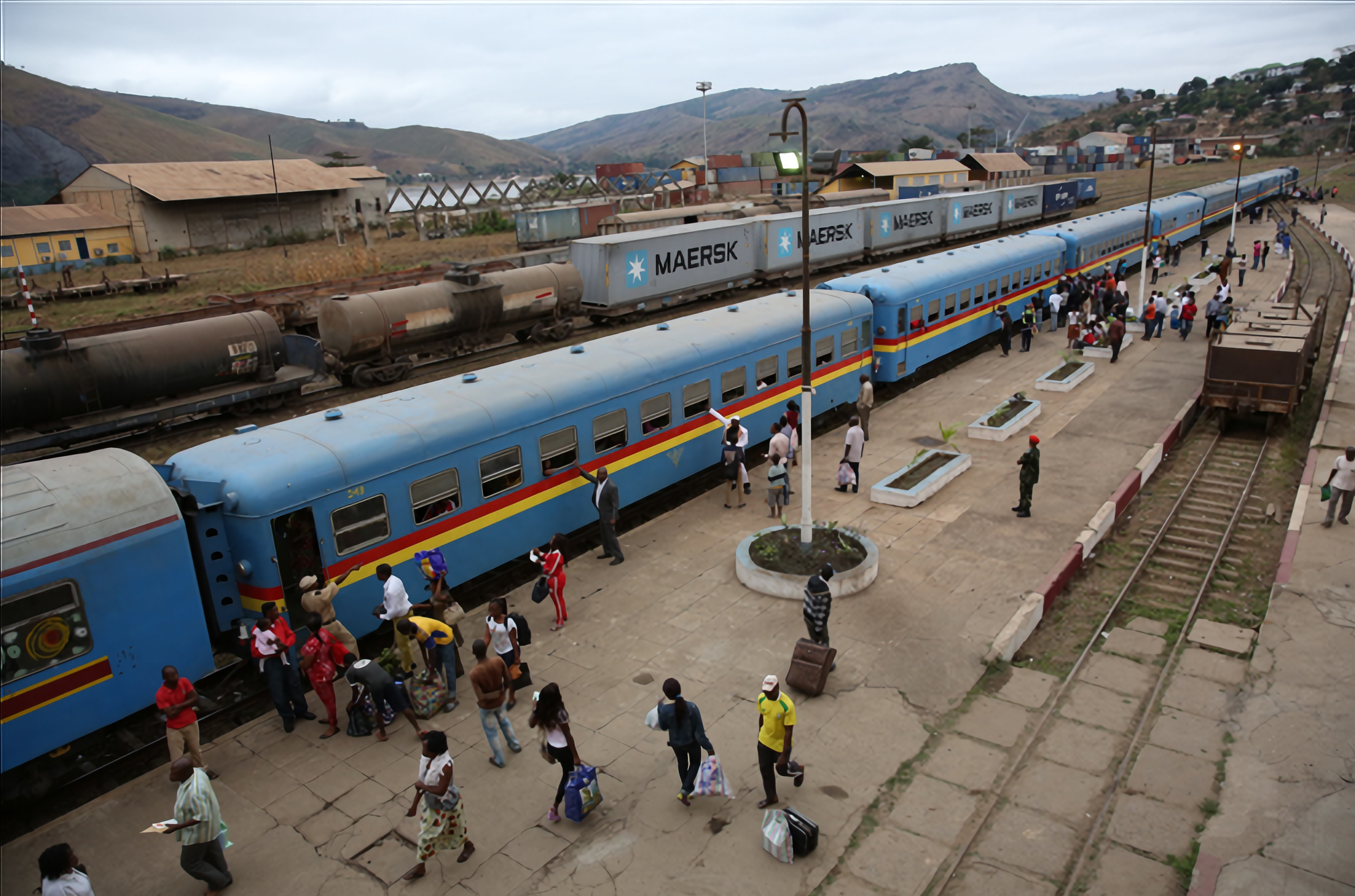}
     \caption{Our method}
    \end{subfigure}
    \caption{The SR result ($\times$4) of VDSR network and our method applied to VDSR network for image DIV2K-0850.}
    \label{fig:0850}
\end{figure*}

\begin{figure*}
    \centering
    \begin{subfigure}[b]{0.49\textwidth}
     \centering
     \includegraphics[width=\textwidth]{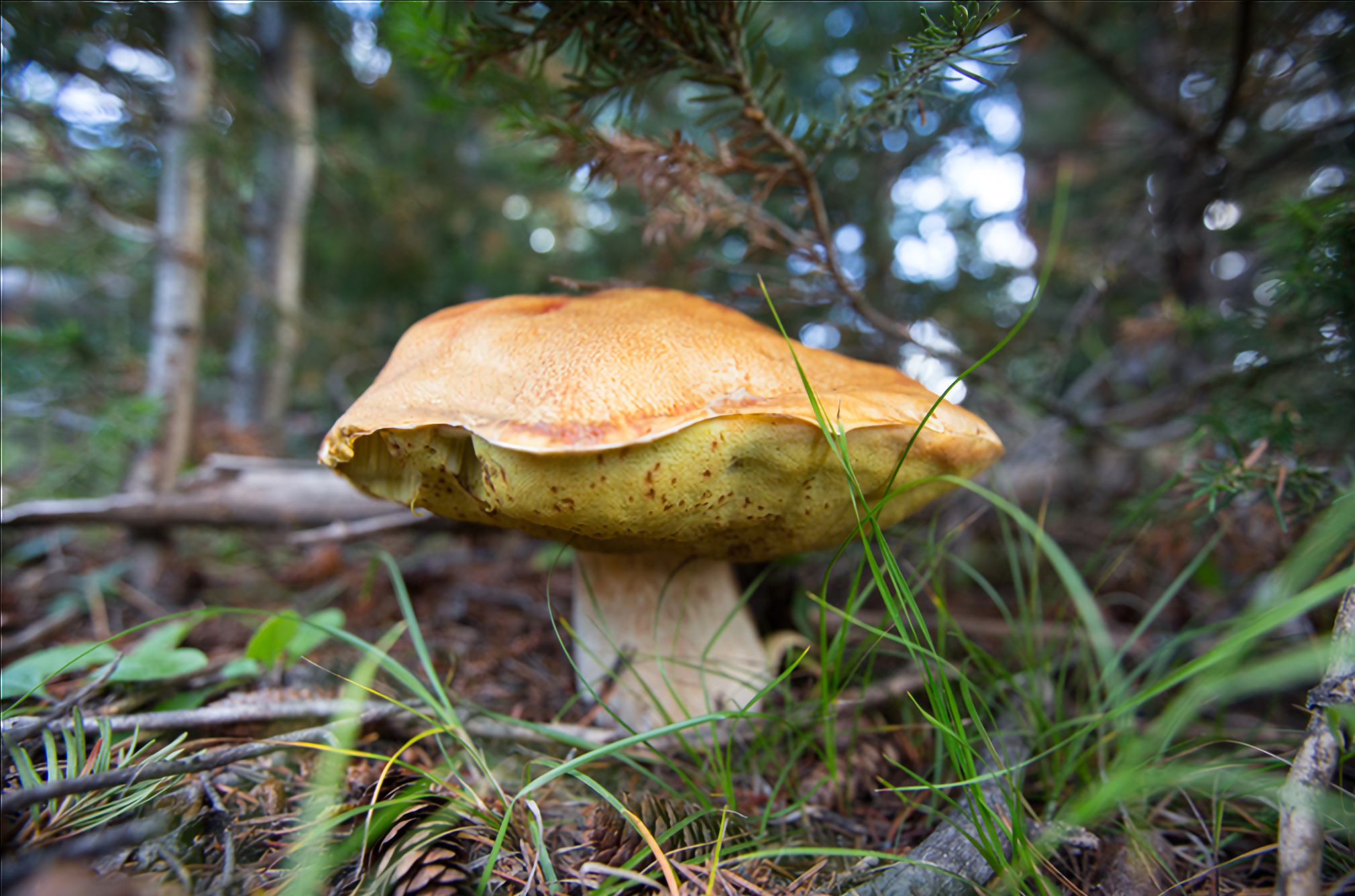}
     \caption{Original Network}
    \end{subfigure}
    \hfill
    \begin{subfigure}[b]{0.49\textwidth}
     \centering
     \includegraphics[width=\textwidth]{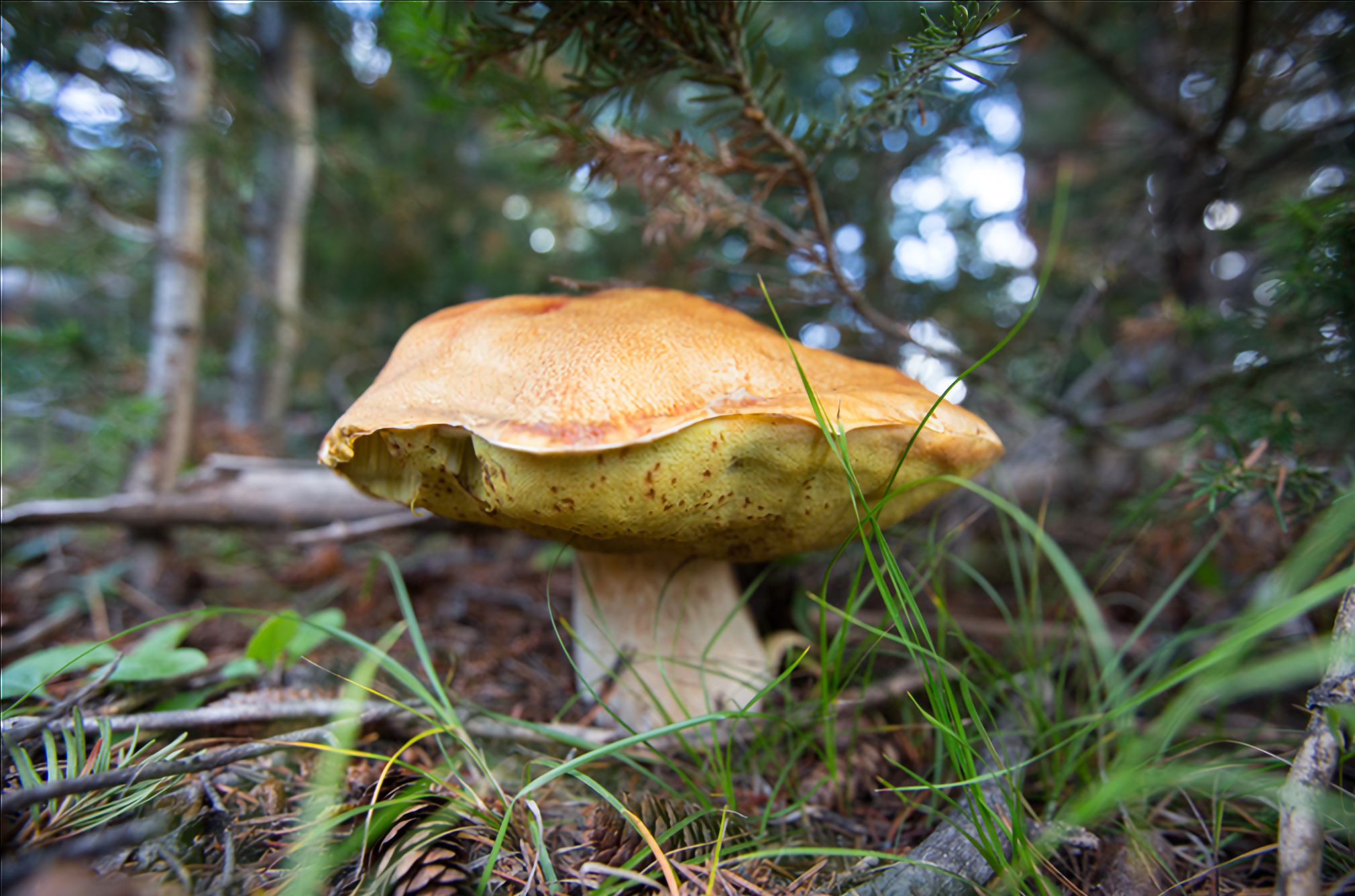}
     \caption{Our method}
    \end{subfigure}
    \caption{The SR result ($\times$4) of VDSR network and our method applied to VDSR network for image DIV2K-0815.}
    \label{fig:0815}
\end{figure*}

This section presents additional qualitative results, demonstrating our perceptual model predictions. \cref{fig:more_maps} shows the maps generated by our method, illustrating the selective deployment of higher-quality reconstruction networks in regions of greater detail and contrast, and the use of lower-quality reconstruction networks in areas with less detail and contrast. \changes{A comparison of the SR results of the proposed method with those of the original networks is presented in \cref{fig:more_images}. 
Furthermore, visual SR results of the proposed method and of the original network is also presented in Fig.~\ref{fig:0885},\ref{fig:0878},\ref{fig:0850},\ref{fig:0815}.}

